# Feature Mapping for Learning Fast and Accurate 3D Pose Inference from Synthetic Images

Mahdi Rad[1]   Markus Oberweger[1]   Vincent Lepetit[2, 1]

[1]Institute for Computer Graphics and Vision, Graz University of Technology, Graz, Austria

[2] Laboratoire Bordelais de Recherche en Informatique, Université de Bordeaux, Bordeaux, France

`{rad, oberweger, lepetit}@icg.tugraz.at`

## Abstract

*We propose a simple and efficient method for exploiting synthetic images when training a Deep Network to predict a 3D pose from an image. The ability of using synthetic images for training a Deep Network is extremely valuable as it is easy to create a virtually infinite training set made of such images, while capturing and annotating real images can be very cumbersome. However, synthetic images do not resemble real images exactly, and using them for training can result in suboptimal performance. It was recently shown that for exemplar-based approaches, it is possible to learn a mapping from the exemplar representations of real images to the exemplar representations of synthetic images. In this paper, we show that this approach is more general, and that a network can also be applied after the mapping to infer a 3D pose: At run-time, given a real image of the target object, we first compute the features for the image, map them to the feature space of synthetic images, and finally use the resulting features as input to another network which predicts the 3D pose. Since this network can be trained very effectively by using synthetic images, it performs very well in practice, and inference is faster and more accurate than with an exemplar-based approach. We demonstrate our approach on the LINEMOD dataset for 3D object pose estimation from color images, and the NYU dataset for 3D hand pose estimation from depth maps. We show that it allows us to outperform the state-of-the-art on both datasets.*

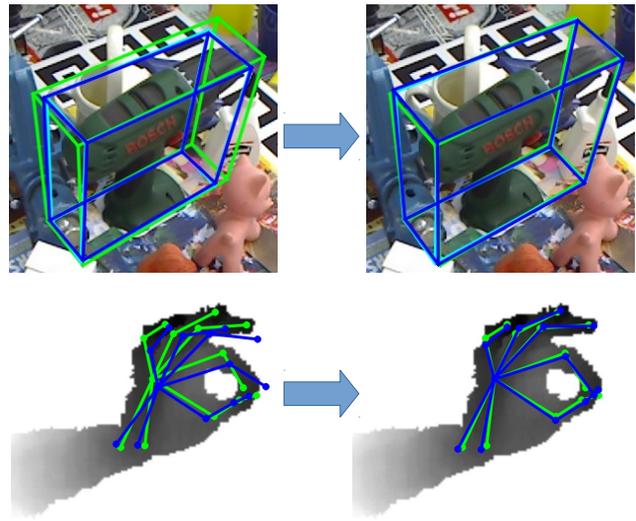

Figure 1: We propose a method for exploiting real and synthetic images to predict a 3D pose from a real image. This method allows us to outperform the state-of-the-art on standard benchmarks. **Top:** The LINEMOD dataset for 3D object pose estimation from color images, and **Bottom:** The NYU dataset for 3D hand pose estimation from depth maps. **Left:** Estimated 3D pose using BB8 [37] and Deep-Prior++ [33] for object and hand, respectively. **Right:** Estimated pose using the method proposed in this paper. Green corresponds to ground truth, blue to our predictions. We obtain the best performances reported so far on these two datasets. (Best viewed in color)

## 1. Introduction

The power of Deep Learning for inference from images has been clearly demonstrated over the past years, however, for many Computer Vision problems, inference is effective only if a large amount of training data is available. Typically this data is created and labelled manually, which is a task expensive in terms of both money and time. Compared to 2D problems where the labels can be directly defined in the training images, the problem is even exacerbated for 3D problems where the training images have to be labelled with 3D data. This 3D data cannot be guessed easily by the human annotator, and needs to be estimated with an *ad hoc* method, for example by using markers [19] or a semi-automatic approach [45].

Many works therefore aimed at using synthetic images created with computer graphics methods [1, 17, 25, 26, 42].

The resulting performances are usually suboptimal, as the synthetic images do not correspond exactly to real images. When some real images are available for training, which is often the case in practice, it is possible to use transfer learning [2, 30, 36, 38], where a first predictor is trained on real images and a second one on synthetic images. By enforcing constraints on the parameters of the two predictors, the first predictor can benefit from a large amount of synthetic training images. Many works using Generative Adversarial Networks (GANs) have also been developed recently [4, 14, 40, 54], in which a first Deep Network is trained to generate real images and it competes with a second Deep Network trained to distinguish synthetic images from real ones.

However, transfer learning and GANs are two general approaches. While important by itself, the 3D pose estimation problem has some specificities that are not exploited by these two approaches. It was shown recently in [29] that, by synthesizing views of objects under the same pose as in some available real images, it is possible to learn a mapping between the features computed for a real image and the features computed in a synthetic image corresponding to the same pose. However, [29] applies this mapping to the descriptors of exemplars, which are matched using special layers computing a similarity score with reference exemplars. In this paper, we show that this mapping can be used as input to a general network. We therefore train a network jointly with the feature mapping to predict the 3D pose of a target object from its synthetic images. We can use a virtually infinite number of training images to train this network, and it therefore performs very well. At run-time, given a real input image of a target object, we compute its image features, map them to the space of features of synthetic images, to finally predict the 3D pose of the object from the mapped features.

As illustrated in Fig. 1, we demonstrate our approach on two different problems: 3D object pose estimation from color images using the LINEMOD dataset [19], and the NYU dataset [45] for 3D hand pose estimation from depth maps. Our experiments show that in both cases, we can significantly outperform the state-of-the-art, by relying on our approach to exploit synthetic images. Moreover, pose inference is very efficient as it is performed by a Deep Network, in contrast to comparisons of exemplars as was done in [29].

In the remainder of this paper, we discuss work related to using synthetic images for training Deep Networks, then present our approach and its evaluation on the LINEMOD and the NYU datasets.

## 2. Related Work

A major problem in training Deep Networks is the acquisition of training data, but training data is critical for the success of Deep Networks [43]. An appealing solution is to use training samples rendered from 3D models [21, 23]. Such annotated samples are very easy to acquire, due to the presence of large scale 3D model datasets [7, 52]. However, using synthetic data requires special measures to prevent the network from overfitting on the synthetic appearance of the data. To prevent overfitting, [21] uses pretrained feature extractors from image classification networks, such as VGG [41], together with sophisticated data augmentation. While this is convenient as no real images are needed, it was only demonstrated on detection problems. [23] also uses synthetically generated images from 3D models with pretrained features, however, they require extensive refinement of the initial network predictions, and we will show that by combining some real images and many synthetic images, we can reach better performances.

Synthetic data on one side and real data on the other side can be seen as two different domains, which gives rise to domain adaptation methods. However, there can be significant differences between the synthetic and real images, which makes methods trained only on synthetic data perform poorly in practice [21]. Domain adaptation techniques provide a welcome solution to this problem, since it is easy to get training data in the synthetic domain, and it can be hard to acquire many training samples in the real domain.

For Deep Networks, fine-tuning is one of the most prominent and simple domain adaptation methods [13, 35]. This, however, can lead to severe overfitting, if there is only a small amount of training labels in the target domain available. Another way to handle the domain shift is to explicitly align the source and target distributions of the data. This can be achieved by quantifying the similarity of the two distributions and maximizing the similarity thereof. One popular metric is Maximum Mean Discrepancy (MMD) [15]. MMD can be either used to align the distributions of target and source features [22, 47], or by learning a transformation of the data such that the distributions match in a common subspace [2, 28, 30, 36]. [12] uses a deep feature extractor together with an additional classifier that predicts the domain for each sample. If the learned features are domain-invariant, such a classifier should exhibit poor performance. [47] adds an MMD loss to align the source and target data representations learned by Deep Networks.

However, [51] observed that feature transferability drops in higher layers of a Deep Network. To leverage this fact, [27] proposed a novel architecture that has the first few layers frozen, the mid layers finetuned, and the fully-connected layers learned for each domain separately. The features of the fully-connected layers are constrained by MMD. This works well for discriminative approaches that separate features into clusters, but not for regression problems. It also requires extensive task-specific validation on which layers to freeze, fine-tune, and transfer.

Very recently, [38] proposed a Siamese Network for do-main adaptation, but instead of sharing the weights between the two streams, their method allows the weights to differ and only regularizes them to keep them related. This method is very general as it can learn adaptation between very different domains. We show in the results section that our approach outperforms [38], as we can exploit the specificities of our problem by rendering synthetic images under the same poses as the real images to learn a mapping.

With the development of Deep Learning, Generative Adversarial Networks (GANs) were also proposed for domain adaptation [14, 54], where a network is trained to transfer images from one domain to another domain. Although GANs are able to generate visually similar images in terms of appearance between different domains [4, 40], the synthesized images lack precision required to train 3D pose estimation methods, as our comparisons to [4] in the results section show. Especially for geometric tasks, such as pose estimation, this shortcoming can be attributed to the lack of geometry in GAN models [50, 54]. An alternative to generating images in order to bridge the domain gap was presented by [46] and [11], who use a domain discriminator with adversarial loss to force a network to learn cross-domain features. Although this works well for discriminative applications, the features are not well suited for regression as we will discuss in the results section.

Another approach for domain adaptation is introduced in [6], which casts domain adaptation as an assignment problem, where samples from the target domain are assigned predefined classes from the source domain. This, however, strongly depends on the initialization of the features to obtain a semantically meaningful mapping, and for regression problems the undefined number of classes would make the problem intractable.

[29] proposed the method most related to ours, as they also learn a mapping from the real to synthetic domain. However, they consider an exemplar-based approach for 3D pose retrieval, and the mapping is applied to the exemplar representations. Relying on exemplars requires a discretization of the pose space in order to create the exemplars, and a nearest neighbor search. A fine discretization thus improves the accuracy of the estimated pose, but also slows down the nearest neighbor search. We show in this paper that it is possible to learn a similar mapping jointly with a network inferring the 3D pose after feature mapping. As our experiments show, our approach of retrieving the pose is thus both fast and accurate.

## 3. Approach

Since it is easy to create synthetic images, our goal is to exploit such images to guide learning, when training a Deep Network to predict a 3D pose from a real image. This real image can be a color image or a depth map.

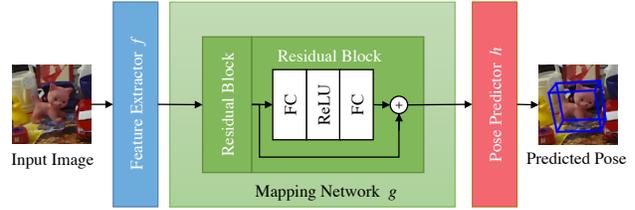

Figure 2: Our model architecture is made of a feature extractor $f$ (blue), a pose predictor $h$ (red), and a network $g$ to map features extracted from a real image to the feature space of synthetic images (green). We apply $g$ only to the features from real images, not on the features from synthetic images. We can thus train the pose predictor on synthetic images, and apply it to real images at run-time without being affected by the domain gap. Within the mapping network, FC denotes a fully-connected layer, and ReLU a rectified linear unit.

### 3.1. Training

We use synthetic images to train a feature extractor $f(\mathbf{x}; \theta_f)$ together with a 3D pose predictor $h(\mathbf{f}; \theta_h)$, which predicts a 3D pose given features $\mathbf{f}$ extracted from a given image $\mathbf{x}$. $\theta_f$ and $\theta_h$ denote the parameters for networks $f$ and $h$, respectively. However, synthetic images do not resemble real images, and their features also vary significantly. Although this might not harm easier tasks such as object detection or object recognition, it is very important for accurate 3D pose prediction.

We therefore train a network $g(\mathbf{f}; \theta_g)$ with parameters $\theta_g$ to map features of the real images into the synthetic feature space, before they are used as input to the $h$ network that predicts the 3D pose from image features. Fig. 2 shows the three networks and how they are connected.

Network $g$ is trained using pairs of images, each pair is made of one of the available real images and of one synthetic image of the target object rendered under the same 3D pose as in the real image. During training, we minimize the distance between the features extracted from the synthetic images and the features extracted from the real image after mapping by $g$. In the case of 3D pose estimation for color images, given a real image, we render the object's 3D model over the real image to obtain the corresponding synthetic images. The first row of Fig. 3 shows an example. In the case of pose estimation from depth maps, we directly generate a depth map of the 3D model under the same pose.

More exactly, we train the three networks $f$, $h$, and $g$ jointly on the training set $\mathcal{T} = \mathcal{T}^{\mathcal{S}} \cup \mathcal{T}^{\mathcal{R}}$ where $\mathcal{T}^{\mathcal{S}} = \{(\mathbf{x}_i^{\mathcal{S}}, \mathbf{y}_i^{\mathcal{S}})\}_i$ denotes a training set of synthetic images and their corresponding 3D labels, and $\mathcal{T}^{\mathcal{R}} = \{(\mathbf{x}_i^{\mathcal{R}}, \mathbf{y}_i^{\mathcal{R}})\}_i$ is made of real images and their 3D labels. We also use a training set $\mathcal{T}^{\mathcal{M}}$ made of the real images $\mathbf{x}_i^{\mathcal{R}}$ in $\mathcal{T}^{\mathcal{R}}$, each paired to a synthetic image generated under the same pose as mentioned above. Jointly training $f$, $h$, and $g$ helps learning to extract image features such that they are transferable from the real domain to the synthetic domain.

We optimize the following loss function over the parameters of the three networks:

$$\mathcal{L}(\theta_f, \theta_h, \theta_g; \mathcal{T}^\mathcal{S}, \mathcal{T}^\mathcal{R}, \mathcal{T}^\mathcal{M}) = \mathcal{L}_{h_S} + \beta \mathcal{L}_{h_R} + \gamma \mathcal{L}_g \,. \quad (1)$$

$\beta$ and $\gamma$ are meta-parameters to control the interaction of the losses.

$\mathcal{L}_{h_S}$ is the loss for predicting the poses for synthetic images:

$$\mathcal{L}_{h_S} = \sum_{(\mathbf{x}_s, \mathbf{y}_s) \in \mathcal{T}^\mathcal{S}} \|h(f(\mathbf{x}_s; \theta_f); \theta_h) - \mathbf{y}_s\|^2 \,. \quad (2)$$

Note that we compose network $f$ that extracts image features from image $\mathbf{x}$ and network $h$ that predicts a 3D pose from these features.

In practice, for parameterizing the 3D pose $\mathbf{y}$ of rigid objects, we use the representation proposed in [37], that is the 2D reprojections of the 3D corners of the object's bounding box, which allows us to use the Euclidean norm in Eq. (2). This representation was shown to be easy to predict by a Deep Network from a color image. Moreover, the 3D pose can be accurately computed from these reprojections using a P$n$P algorithm. In the case of the hand pose estimation, $\mathbf{y}$ is simply a vector made of the 3D locations of the hand joints normalized as in [31].

$\mathcal{L}_{h_R}$ is a loss function equivalent to $\mathcal{L}_{h_S}$ but for the real images:

$$\mathcal{L}_{h_R} = \sum_{(\mathbf{x}_r, \mathbf{y}_r) \in \mathcal{T}^\mathcal{R}} \|h(g(f(\mathbf{x}_r; \theta_f); \theta_g); \theta_h) - \mathbf{y}_r\|^2 \,, \quad (3)$$

where we compose $f$, $g$, and $h$ together, to first extract image features, then map them to the space of image features for synthetic images, and finally predict the 3D pose from these mapped features.

$\mathcal{L}_g$ is the loss to learn the mapping between the features extracted from the real images to the features extracted from the synthetic images:

$$\mathcal{L}_g = \sum_{(\mathbf{x}_r, \mathbf{x}_s) \in \mathcal{T}^\mathcal{M}} \|g(f(\mathbf{x}_r; \theta_f); \theta_g) - f(\mathbf{x}_s; \theta_f)\|^2 \,. \quad (4)$$

### 3.2. Effect of the Learned Mapping

To understand better the effect of the learned mapping, we computed the distributions of the absolute differences between the synthetic feature vectors and real feature vectors, as computed by network $f$, before and after mapping of the real feature vectors by network $g$. The distributions remain surprisingly close, as their means and standard deviations are $(\mu_1 = 1.60, \sigma_1 = 1.64)$ and $(\mu_2 = 1.30, \sigma_2 = 1.50)$ respectively.

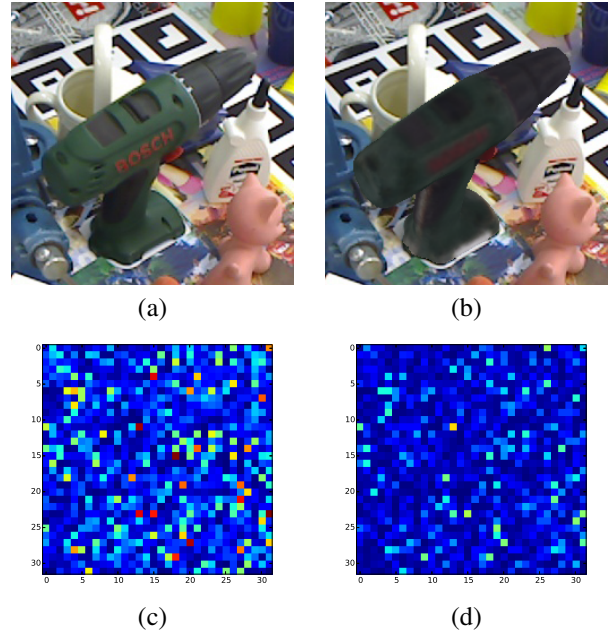

Figure 3: The effect of the mapping learned by network $g$ using a real image (a) and the corresponding synthetic image (b). The second row shows the absolute differences between the synthetic and real feature vectors (reshaped to images for better visualization) before (c) and after mapping (d). The mapping mostly removes the large differences.

However, considering distributions can only provide a limited view. To get a finer insight, we took a pair of real and synthetic images under the same pose, and we plotted the absolute differences between the coefficients of their feature vectors, first without mapping, then after mapping of the real feature vector by network $g$. The differences are shown in Fig. 3(c) and Fig. 3(d). Blue corresponds to small differences, and red to large differences. It appears that the mapping mostly removes the large differences without changing the smaller differences. We repeated this experiment on other image pairs and observed a similar behavior. Our interpretation is that for each pair, only a few feature coefficients are responsible for the domain gap, and they can be attenuated by the mapping.

### 3.3. Pose Prediction

At run-time, given a real image $\mathbf{x}$, we predict the corresponding 3D pose $\mathbf{y}$ by composing the three networks together:

$$\mathbf{y} \leftarrow h(g(f(\mathbf{x}; \widehat{\theta}_f); \widehat{\theta}_g); \widehat{\theta}_h) \,,$$

where $(\widehat{\theta}_f, \widehat{\theta}_g, \widehat{\theta}_h)$ are the networks' parameters found during training. Note that this composition of the three networks can be implemented as a single network to improve efficiency.

### 3.4. Network Details

As shown in Fig. 2, we use two Residual blocks [18] for network $g$ to map feature vectors of size 1024. Each fully-connected layer within the Residual block has 1024 neurons.

The network architectures of $f$ and $h$ depend on the application. In the case of 3D object pose estimation from color images, we use the VGG-16 network [41] for initializing $f$, where we use the first 10 convolutional layers as feature extractor, and we add two fully-connected layers with 1024 neurons, as in [37]. For $h$ we use a single fully-connected layer with 16 outputs—2 coordinates for each corner of the bounding box.

In the case of 3D hand pose estimation from depth maps, we use an architecture similar to the one of [31, 33] for the feature extraction network $f$: It is similar to the 50-layer Residual Network [18] with 4 residual modules. We remove the Global Average Pooling [18], which we experienced during our experiments significantly reduces localization accuracy of the joints, and add two fully-connected layers with 1024 neurons each. $f$ is trained from scratch. We use a single fully-connected layer with 42 outputs—3 for each of the 14 joints—for the pose prediction network $h$.

For the parameters of the loss function in Eq. (1), we use $\beta = 1$, which gives the same weight to the synthetic and real samples, and $\gamma = 0.2$, which gives a good trade-off between pose loss and feature mapping loss. We optimize the loss using gradient descent, specifically the ADAM algorithm [24]. In order to improve convergence, we pretrained the networks with synthetic data only, which in practice gives better results compared to starting from a random initialization.

## 4. Experiments

In this section, we present and discuss the results of our evaluation on two different applications, *i.e.* 3D object pose estimation from RGB images and 3D hand pose estimation from depth maps.

### 4.1. 3D Object Pose Estimation

We first compare our method to other domain adaptation methods: The transfer learning method proposed in [38], the GRL method of [12], and the ADDA method of [46], on 3D object pose estimation using the LINEMOD benchmark [19]. We also compare to the DDC method [47], designed for classification. We therefore changed the classification loss to our regression loss on the 3D pose. For all these methods, we use the VGG-16 network [41] as described in Section 3.4, which is also the network used in [37].

**Training set creation.** For creating the training set $\mathcal{T}^S$ of Eq. (2), we generate in total 5M synthetic training images online during training, from poses randomly sampled on the upper hemisphere of the object. The in-plane rotation is randomly sampled within range $[-45°, +45°]$, scale within range $[65cm, 115cm]$. We use the 3D models provided with the LINEMOD dataset, and render them over random backgrounds extracted from images of ImageNet [8]. We do not simulate lighting as the object textures already exhibit lighting effects. Note that this lighting does not correspond to the lighting of the real images in general. We also do not add blur, nor Gaussian noise, nor any other techniques often applied to synthetic images created for training [21, 38]. For generating the training set $\mathcal{T}^R$ of Eq. (3) we use the same protocol as in [37]: We slightly rescale the image of the segmented object, superimpose it on a random background picked from the ImageNet dataset [8], after a small random translation from the center of the image window.

**Comparison results.** Table 1 provides the 2D Projection metric [5] obtained by these methods using the ground truth 2D object center. Our method significantly outperforms the other domain adaptation methods on this problem. This illustrates that our feature mapping can generalize better to the 3D pose estimation problem compared to existing domain adaptation techniques.

Table 2 provides the final results obtained after the refinement stage of [37] trained with synthetic images using our feature mapping method. We compare them with the results obtained by current state-of-the-art methods, *i.e.* [5], BB8 with refinement [37], SSD-6D [23]. For our method, we use the detection technique of [37].

More precisely, Table 2 reports the measures commonly used on the LINEMOD dataset: The 2D Projection [5], the 6D pose [20], and the 5cm 5° [39] metrics for all methods, except [23] who provides the 6D pose metric only. To the best of our knowledge, our approach obtains the best results obtained on LINEMOD from RGB images reported so far. To appreciate the quality of ours results, we show in Fig. 5 the bounding boxes we retrieve on the same images as in the third column of Figures 4 and 5 of the supplementary material of [23] for comparison where they use RGB refinement. Note that for the comparison with [37], we use our reimplementation, which achieves slightly better results than the numbers reported in [37]. *Training [37] on a large number of synthetic images, together with real images, performs similar to BB8. Training only on synthetic images performs extremely poorly with a performance of 12% for the 2D Projection metric.*

[4] does not report the metrics mentioned above for its GAN approach on the LINEMOD dataset. It does report a mean rotation error of $23.5°$, while our method obtains a significantly lower error of $3.5°$. This indicates that the

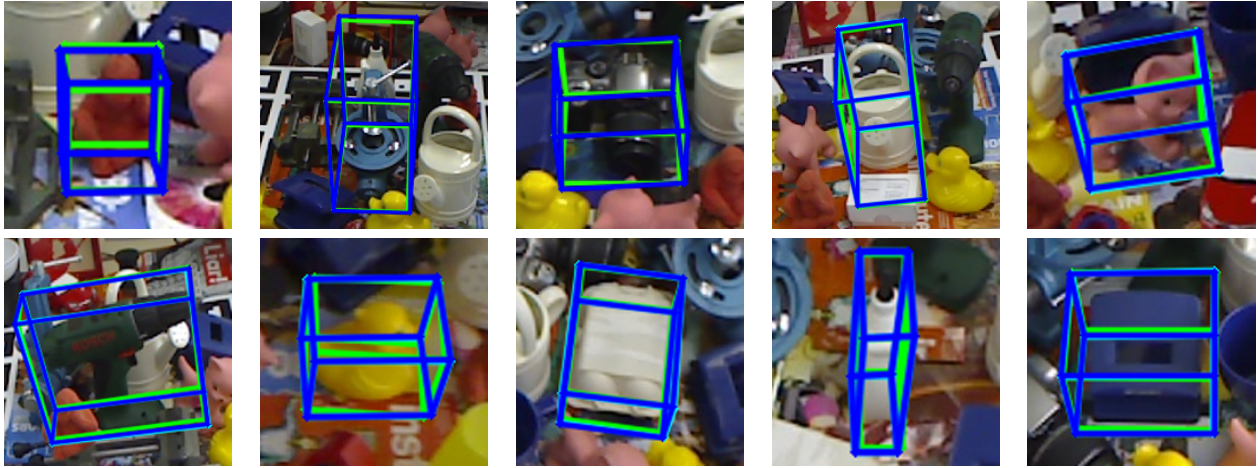

Figure 5: Qualitative results obtained by our method on the LINEMOD dataset [19]. The green bounding boxes correspond to the ground truth poses, and the blue bounding boxes to the estimated poses. Our predictions remain always very close to the ground truth.

| Object | [38] | GRL [12] | DDC [47] | BB8 [37] w/o ref. | Ours |
|---|---|---|---|---|---|
| Ape | 95.7 | 96.0 | 94.4 | 94.0 | **96.6** |
| Bench Vise | 94.6 | 94.8 | 91.3 | 90.0 | **96.3** |
| Camera | 88.6 | 90.4 | 84.0 | 81.7 | **94.8** |
| Can | 95.8 | 95.7 | 95.1 | 94.2 | **96.6** |
| Cat | 96.1 | 97.0 | 95.6 | 94.7 | **98.0** |
| Driller | 75.4 | 78.0 | 70.7 | 64.7 | **83.3** |
| Duck | 94.7 | 95.2 | 93.1 | 94.4 | **96.3** |
| Egg Box | 95.8 | 95.6 | 95.1 | 93.5 | **96.1** |
| Glue | 95.7 | 95.8 | 94.8 | 94.8 | **96.9** |
| Hole Puncher | 91.5 | 91.9 | 92.6 | 87.2 | **95.7** |
| Iron | 87.5 | 88.8 | 83.2 | 81.0 | **92.3** |
| Lamp | 80.5 | 80.8 | 77.7 | 76.2 | **83.5** |
| Phone | 78.3 | 80.9 | 76.0 | 70.6 | **88.2** |
| Average | 90.0 | 90.8 | 88.0 | 85.9 | **93.4** |

Table 1: Comparison of different domain adaptation methods ([38], GRL [12], and DDC [47]) on 3D object pose estimation, using the LINEMOD dataset and the 2D Projection metric [5]. All methods use the ground truth 2D object center and predict the 3D object pose using the 2D projections of the objects' 3D bounding box [37]. BB8 w/o ref. [37] denotes the results obtained using only the available real images. Our method outperforms the other domain adaptation methods on this problem.

| Metric | [5] with ref. | BB8 [37] with ref. | SSD-6D [23] with ref. | Ours with ref. |
|---|---|---|---|---|
| 2D Projection | 73.7 | 91.8 | - | **95.4** |
| 6D pose | 50.2 | 70.1 | 76.6 | **78.7** |
| 5cm 5° | 40.6 | 73.3 | - | **80.1** |

Table 2: Comparison of our final results to state-of-the-art methods on the LINEMOD dataset. Here [5] and [23] use their own pose refinement methods, while we use the same refinement stage as [37] but trained using our approach, To the best of our knowledge, our approach obtains the best results obtained on LINEMOD from RGB images reported so far.

images generated by the GAN are too inaccurate to infer an accurate 3D pose. We also tried the method of [46], which performed actually more poorly on this problem. This can be explained by the fact that this method does not guarantee that the learned features carry enough information for predicting a pose, as this method was designed for classification, not regression.

**Influence of the number of real images.** Capturing real data is very cumbersome and time consuming, which motivates the use of synthetic data. On the LINEMOD dataset, [5, 37] both use 15% of the images for training, which represents 180 real images per object on average, and the remaining images for testing. Fig. 4 shows the influence of the number of real training images on the final results. Using our method systematically improves the results compared to using only real images for training. We require about half the amount of training images to achieve the same accuracy as [37].

### 4.2. 3D Hand Pose Estimation

To show the generality of our approach, we also evaluate it on 3D hand pose estimation from single view depth maps.

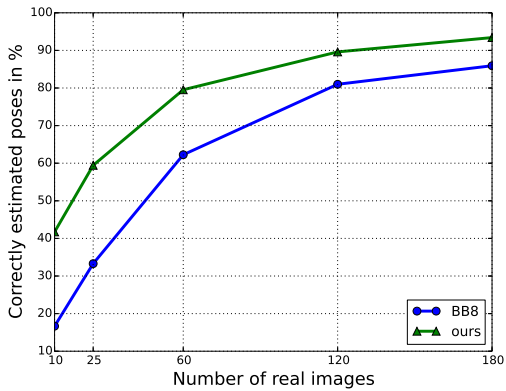

Figure 4: Influence of the number of real images for 3D object pose estimation using the 2D Projection metric [5]. BB8 corresponds to the results obtained when only real images are used for training [37]. When using the same number of training images, our method systematically improves the performance. It needs half the number of images to reach the same performance.

For this experiment, we consider the NYU dataset [45]. This dataset is very challenging as it contains hands from multiple subjects under a large range of 3D poses. Moreover, as can be seen in Fig. 6, it exhibits severe noise due to the use of a structured light sensor to capture the depth maps. The dataset is made of 72k depth maps for training and 8k for testing. The depth maps are captured from three different viewpoints and annotated with the 3D joint locations. We follow the protocol of [45] and predict a subset of 14 joints. We use the pipeline provided by [34] to preprocess the depth maps: It crops a $128 \times 128$ patch around the hand location, and normalizes its depth values to the range of $[-1, +1]$.

We compare our method to very recent state-of-the-art methods: *DeepPrior++* [33] integrates a prior on the 3D hand poses into a Deep Network; *REN* [16] relies on an ensemble of Deep Networks, each operating on a region of the input image; *Lie-X* [49] uses a sophisticated tracking algorithm constrained to the Lie group; *Crossing Nets* [48] uses an adversarial training architecture; Neverova *et al*. [32] proposed a semi-supervised approach that incorporates a semantic segmentation of the hand; *DeepModel* [53] integrates a 3D hand model into a Deep Network; *DISCO* [3] learns the posterior distribution of hand poses; *Feedback* [34] uses an additional Deep Network to improve results of an initial prediction; *Hand3D* [9] uses a volumetric CNN to process a point cloud.

**Training set creation.** Creating synthetic depth maps for hand poses is a relatively simple problem. In practice, we use the 3D hand model of [45] to render synthetic views of a hand. However, it should be noted that the noise present in real depth maps captured with a structured light sensor is difficult to simulate, and our synthetic depth maps do not contain any noise. We use 5M synthetic images of the hand that are rendered online during training from poses of the training set perturbed with randomly added articulations.

| Method | Average 3D error |
|---|---|
| Neverova *et al*. [32] | 14.9mm |
| Crossing Nets [48] | 15.5mm |
| Lie-X [49] | 14.5mm |
| REN [16] | 13.4mm |
| DeepPrior++ [33] | 12.3mm |
| Feedback [34] | 16.2mm |
| Hand3D [9] | 17.6mm |
| DISCO [3] | 20.7mm |
| DeepModel [53] | 16.9mm |
| Synthetic only | 21.1mm |
| **Ours** | **7.4mm** |

Table 3: Quantitative evaluation on the NYU dataset [45]. We compare the average Euclidean 3D error of the predicted poses with state-of-the-art methods on the NYU dataset. The numbers are reported for the real test images.

**Comparison results.** Table 3 compares the different methods we consider using the average Euclidean distance between ground truth and predicted joint 3D locations, which is a *de facto* standard for this problem. When training on synthetic data only, the error on the real test images is 21mm, which suggests that the network severely overfits to the rendered depth maps and cannot generalizes to real depth maps, which are often very noisy. By using our feature mapping, we achieve an error of 7.4mm, which improves the state-of-the-art by 4.9mm or almost 40%.

Fig. 6 shows some qualitative results. When only synthetic depth maps are used, the predictions for synthetic depth maps are typically very good, but the predictions for real frames are bad. When using our method, the predicted poses significantly improve, especially in the presence of noise in the depth maps.

Fig. 7 shows the fraction of frames where all joints of a frame are within a maximum distance from the ground truth. This is a very difficult metric since a single erroneous joint can deteriorate the result of a frame [34, 44]. We significantly outperform all previous works on this difficult metric, by almost 40% at an error threshold of 20mm.

### 4.3. Computation Times

We implemented our approach in Tensorflow [10] and the code runs on an Intel Core i7 3.30GHz desktop with a Geforce TITAN X. For 3D object pose estimation, our reimplementation of [37], on which our approach is built on, is significantly faster than the original implementation: It takes 3.2ms for the pose inference including feature mapping, which requires only a few matrix multiplications in addition to the feature extraction. This corresponds to approximately 300fps. For 3D hand pose estimation, our approach takes 8.6ms, which corresponds to more than 110fps. These computation times should be compared to those of the exemplar-based approach of [29], which reports run-times of several seconds per image.

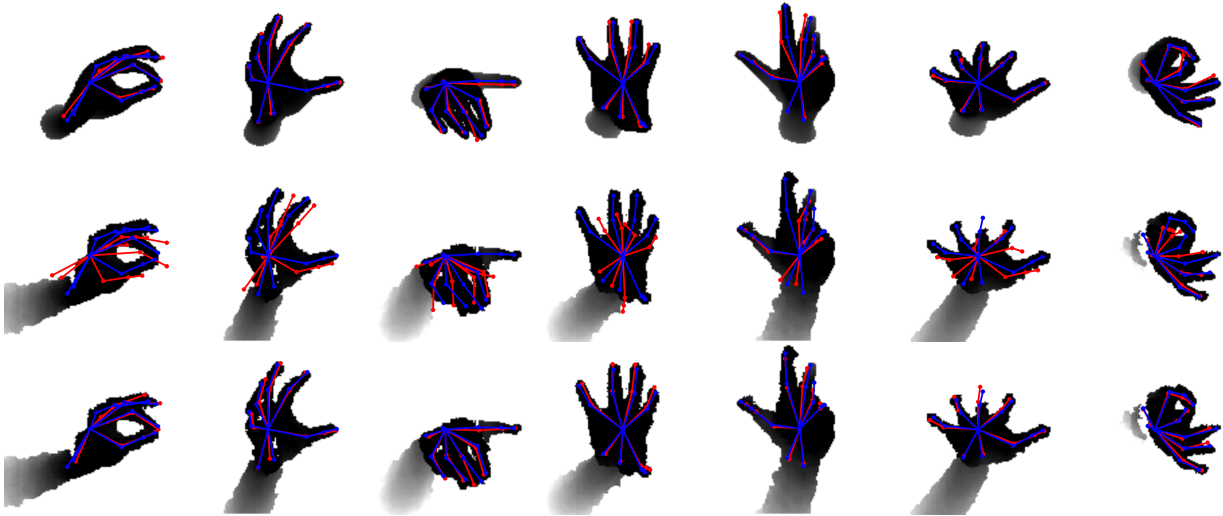

Figure 6: Qualitative results on the NYU dataset [45]. **Top row**: Predicted poses on synthetic test depth maps, and **Middle row**: predicted poses on real test depth maps, when training on synthetic depth maps only. Training only on synthetic depth maps does not generalize well to real depth maps. **Bottom row**: Predicted poses when using our method. We plot the 2D projection of the 3D pose prediction. Blue denotes ground truth annotations, red are our predictions.

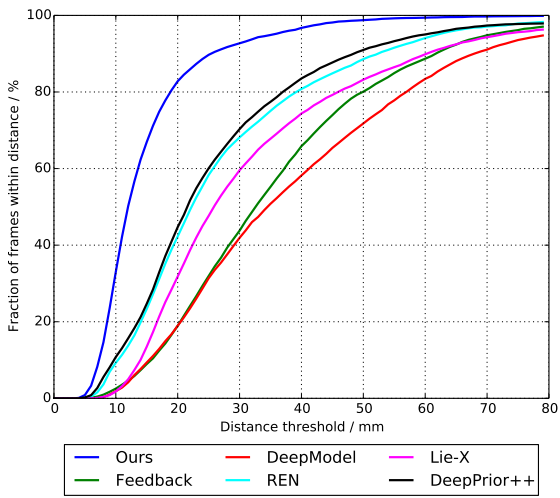

Figure 7: Comparison with state-of-the-art methods on the NYU dataset [45]. We plot the fraction of frames where all joints of a frame are within a maximum distance from the ground truth. A larger area under the curve indicates better results. Our proposed approach performs best among all other methods.

## 5. Conclusion

We showed that domain transfer between synthetic and real images can be achieved easily in the case of 3D pose inference. We presented an approach that learns a mapping between the two domains from synthetic images rendered under the same pose as the available real training images, jointly with feature extraction and pose inference. Our method is simple to implement, can be easily optimized, is very efficient at run-time, and allowed us to outperform the state-of-the-art on popular datasets for 3D object pose and 3D hand pose estimation.

## Acknowledgment

This work was funded by the Christian Doppler Laboratory for Semantic 3D Computer Vision. We thank Alexander Grabner for helpful discussions. We would also like to thank the anonymous reviewers for their work and their constructive comments, in particular Reviewer #1 and the MetaReviewer, and the authors of [23] for their responsiveness during the paper writing period.

# Feature Mapping for Learning Fast and Accurate 3D Pose Inference from Synthetic Images - Supplementary Material

## 6. Comparison to State-of-the-Art Baselines

We compare against BB8 [37] and DeepPrior++ [33], which are the second best performing methods identified in the main paper. Fig. 8 shows a detailed numerical comparison with these methods on the LINEMOD dataset and the NYU dataset.

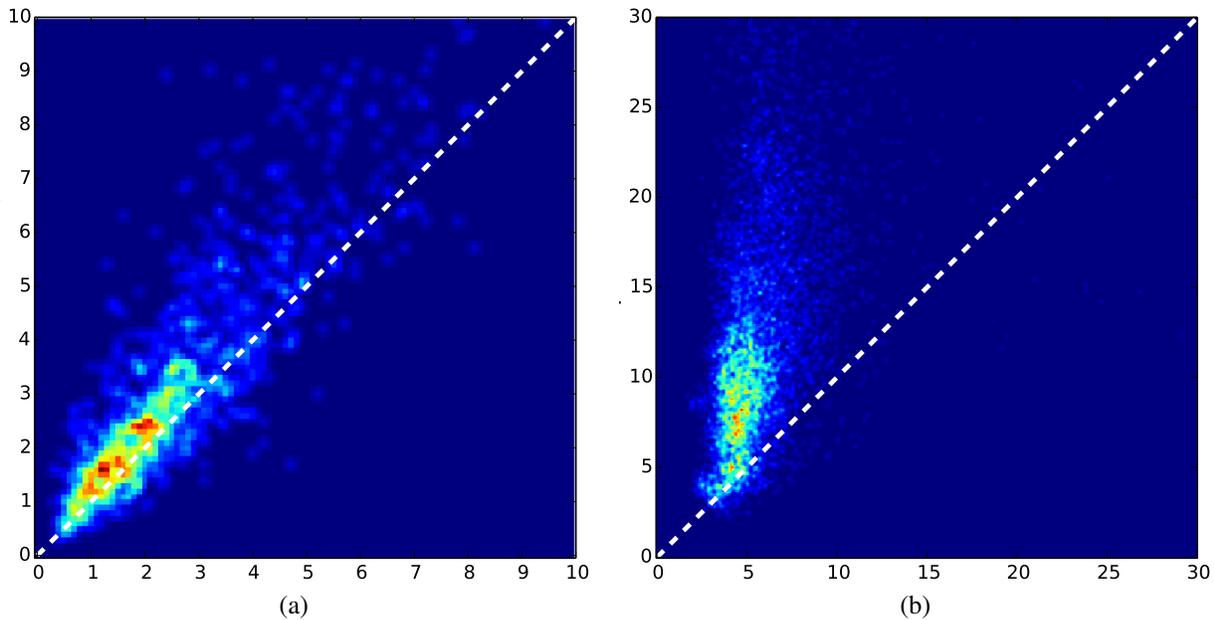

Figure 8: (a) Joint distribution of the 2D projection metrics (in pixel) on the LINEMOD dataset for our approach (on the x-axis) and BB8 [37] (on the y-axis), both using the RGB refinement of [37]. (b) Same for the mean Euclidean distance (in mm) on the NYU dataset for our approach and DeepPrior++ [33]. The fact that the distribution mass is over the main diagonal shows that our approach improves the pose estimates for most of the images. The improvement is often very large on NYU, and we can still improve the performance on the LINEMOD dataset on which BB8 already performs very well.

## 7. Qualitative Results

Fig. 9 shows how much our approach can improve the results of BB8 on the LINEMOD dataset. When a similar pose is present in the training set, we can still often keep improving the accuracy with respect to BB8, as shown in Fig. 10. Fig 11 shows that our estimated poses are accurate enough for Augmented Reality applications.

Figs. 12 and 13 show 3D hand pose estimates obtained with our approach compared to those obtained with DeepPrior++.

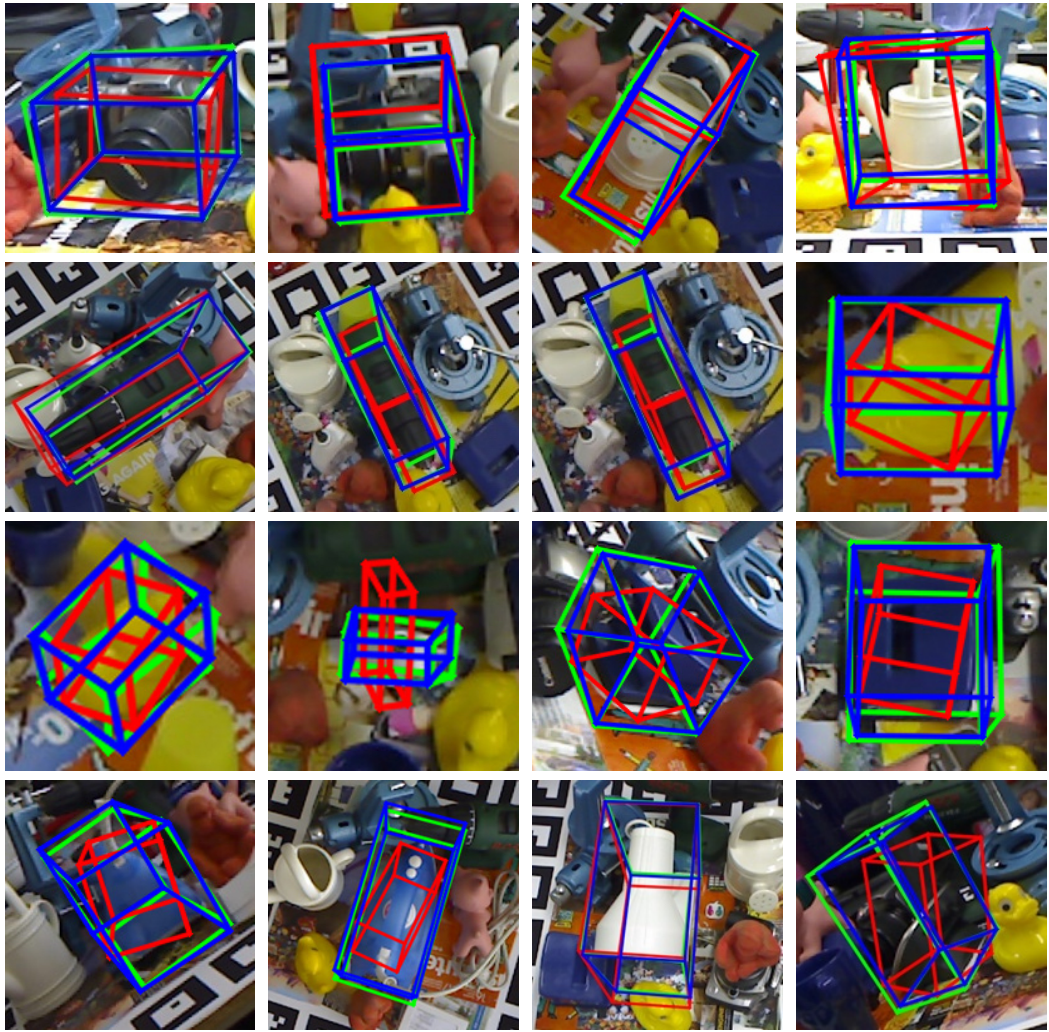

Figure 9: Zooms on pose estimated on the LINEMOD dataset [19], using our approach and the RGB refinement of [37] also trained using our method, where BB8 [37] fails mostly due to the lack of corresponding poses in the training set. The green bounding boxes correspond to the ground truth poses, and the red and blue bounding boxes to the estimated poses using BB8 and our approach, respectively.

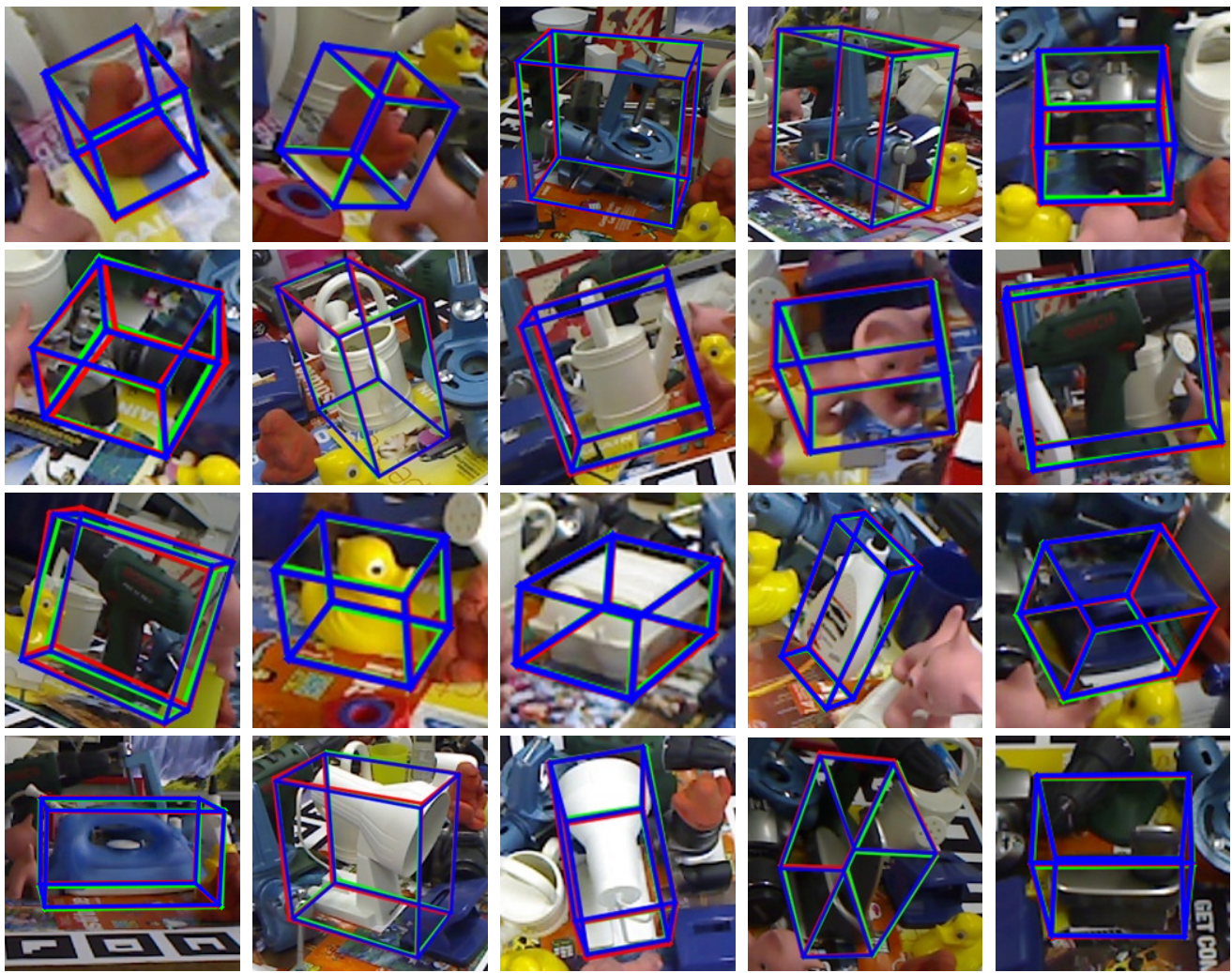

Figure 10: Zooms on pose estimated on the LINEMOD dataset [19], using our approach and the RGB refinement of [37] also trained using our method. When a similar pose is present in the training set, as it is the case here, we can still often keep improving the accuracy with respect to BB8. The green bounding boxes correspond to the ground truth poses, and the red and blue bounding boxes to the estimated poses using BB8 and our approach, respectively.

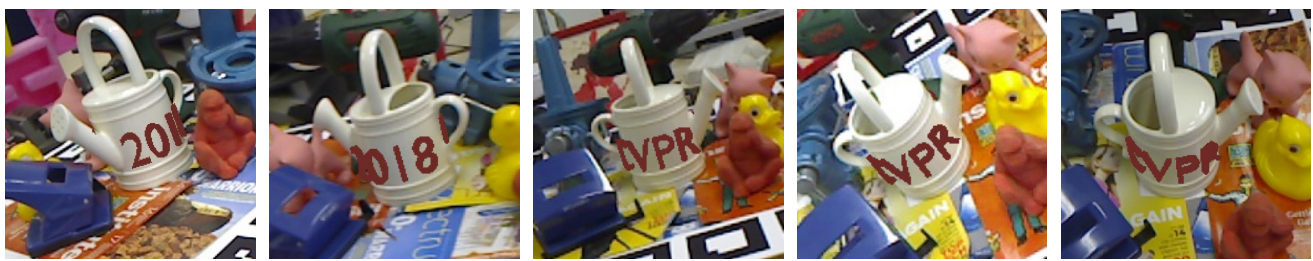

Figure 11: Our poses are accurate enough for Augmented Reality applications. Given our estimated poses, we render "CVPR 2018" on the Can of the LINEMOD dataset.

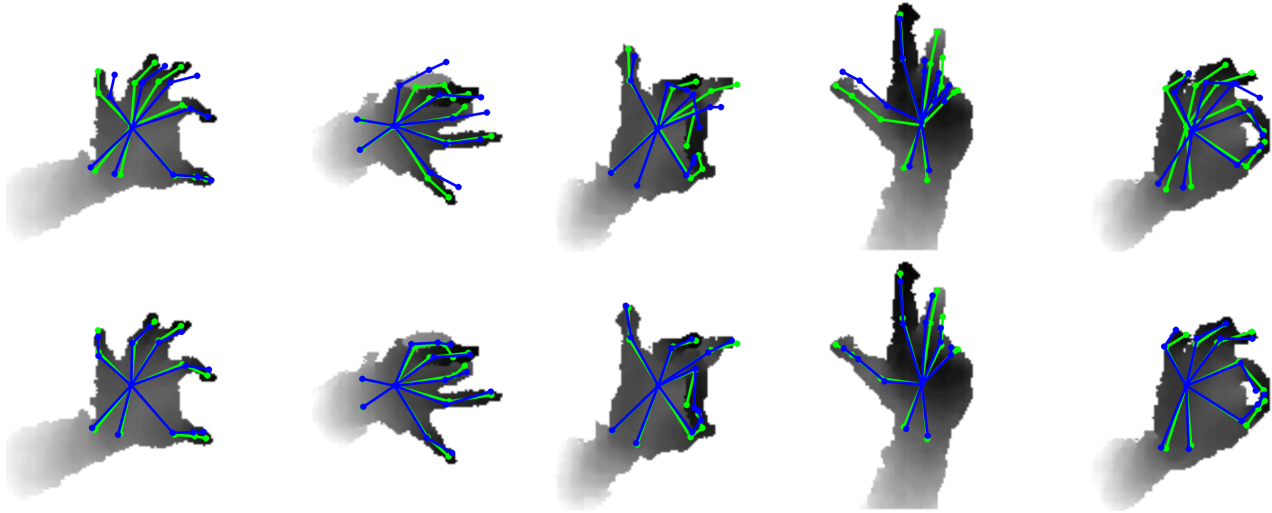

Figure 12: **Top row:** Estimated 3D poses for the hand on the NYU dataset [45] by using the state-of-the-art method DeepPrior++ [33]. **Bottom row:** estimated poses using our approach. Green corresponds to ground truth, blue to our predictions.

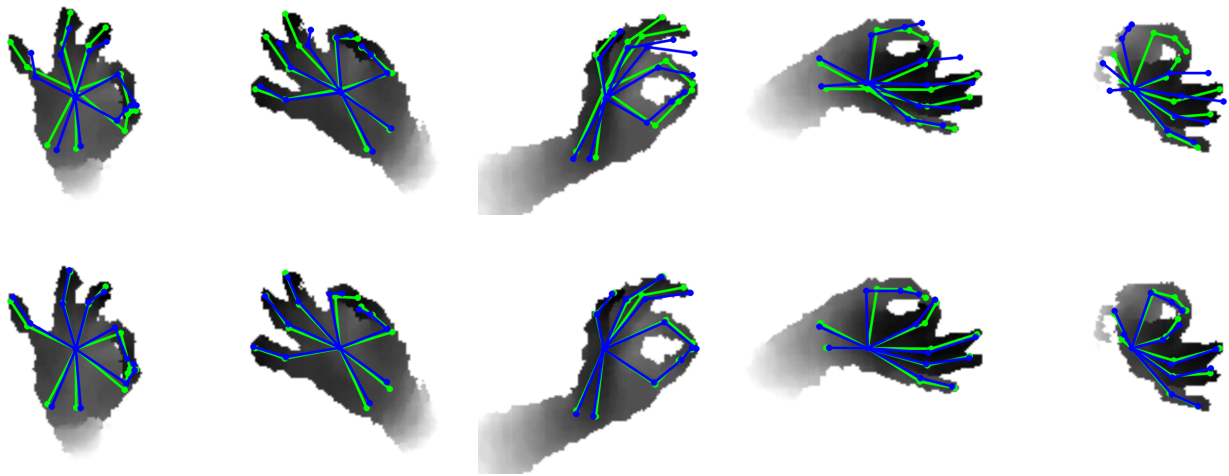

Figure 13: **Top row:** estimated 3D poses for the hand on the NYU dataset [45] by using the state-of-the-art method DeepPrior++ [33]. **Bottom row:** estimated poses using our approach. Green corresponds to ground truth, blue to our predictions.